# Artificial Intelligence Based Predictive Maintenance for Electric Buses

[1]Ayşe Irmak Erçevik[0009-0000-0974-2527] and [2]Ahmet Murat Özbayoğlu [0000−0001−7998−5735]

[1]TOBB University of Economics and Technology, Faculty of Engineering, Department of Computer Engineering, 06510 Ankara, Turkey
`ayseirmakercevik@etu.edu.tr,`
[2] TOBB University of Economics and Technology, Faculty of Engineering, Department of Computer Engineering, 06510 Ankara, Turkey
`mozbayoglu@etu.edu.tr`

**Abstract**

Predictive maintenance (PdM) is crucial for optimizing the efficiency and minimizing downtime of electric buses. While these vehicles provide significant environmental benefits, they present challenges for PdM due to the complexity of the electric transmission and battery systems. Traditional maintenance approaches, often based on scheduled inspections, struggle to capture anomalies in the real-time CAN BUS data of electric buses which are multi-dimensional, and often complex to interpret. This paper is focusing on employing a graph-based feature selection method to analyse the relationships between the characteristics of a CAN (Controller Area Network) Bus dataset pertaining to electric buses. Furthermore, it investigates the prediction performance of targeted alarms utilising artificial intelligence techniques. The study employed data obtained from CAN Bus systems of diverse vehicles over an extended period. The raw data packages underwent extensive pre-processing techniques to create datasets suitable for the structural needs of machine learning models. In this process, a hybrid graph-based feature selection tool was developed using a combination of statistical filtering methods, including Pearson correlation analysis, Cramer's V statistical table and ANOVA F-test, and optimisation-based community detection algorithms, such as InfoMap, Leiden, Louvain and Fast Greedy. The developed tool was employed to identify feature subsets, evaluate the relational position of the features and their relationship with the targeted alarms, select distinctive features and perform dimension reduction. In the selection of machine learning models, a number of different approaches like the Support Vector Machine (SVM), Random Forest (RF) and eXtreme Gradient Boosting (XGBoost) classifiers were applied, and their performance was then evaluated. In order to mitigate the impact of class imbalance on model performance, SMOTEEN (Synthetic Minority Oversampling Technique-Edited Nearest Neighbours) and binary search-based time interval down-sampling methods were employed. To identify the optimal distinctive classifier space for each model, a grid search and random search-based hyperparameter optimisation (hyperparameter tuning) was conducted. The performance of the developed models was evaluated based on the metrics employed for model validation. The predictions of the models were explained by examining the weights (importance) of the features and utilising the LIME (Local Interpretable Model-Agnostic Explanations) tool to identify the features that triggered the predictions, thereby gaining insights into the functionality (design) of the system. The present study has investigated the prediction performance of specific alarms in vehicles utilising CAN Bus data with machine learning techniques. Additionally, a graph-based feature selection tool has been developed, enabling the visual interpretation of relationships between features and facilitating the enhancement of corrective, preventive, predictive and proactive maintenance (design improvement) approaches. It is concluded that the techniques employed throughout the machine learning life cycle, as conducted in the study, can provide valuable insights and contribute to the advancement of maintenance strategies through machine learning, particularly in contexts where data quality and access to expert opinion may present challenges, in alignment with the principles of Industry 4.0.

# 1    Introduction

This study investigated the predictive performance of fault/warnning alarms in electrical vehicles using machine learning techniques on CAN bus data and developed a graph-based feature selection tool to provide a visual interpretation of the relationships between features to help to improve corrective, preventive, predictive and proactive maintenance approaches.

Mobley states that maintenance costs contribute a significant proportion of total production costs, ranging from 15% to 60% depending on the sector (Mobley, 2002). It is also important for bus operators to consider maintenance as a vital element of fleet management (Raposo, Farinha, Ferreira, & Galar, 2017) (Ambriško & Teplická, 2021) as it has a significant impact on various aspects of fleet operations. Maintenance plays a crucial role in ensuring energy efficiency, controlling carbon emissions, protecting the value of assets (Villar, 2024), and ultimately, customer satisfaction. In the 4.0 era, the optimisation of maintenance strategies through the utilisation of machine learning methodologies has been demonstrated to be an effective approach for the reduction of costs and the minimisation of service interruptions (Theissler, Pérez-Velázquez, Kettelgerdes, & Elger, 2021) (Cinieri, 2019) The aim of this study is to apply machine learning techniques and feature analysis to CAN Bus data obtained from electric buses in order to predict alarms associated with potential faults in vehicles. Furthermore, electric buses are complex systems that contain a large number of components and sensors and generate an extensive amount of data. This presents a significant challenge in understanding the interrelationships between components and identifying the components associated with fault alarms (Hossan & Chowdhury, 2019). This study addresses the complexity of the dataset by employing advanced feature selection and model explainability techniques. In this manner, not only fault alarms are predicted, but also valuable insights are derived regarding the performance and structure of the system by analysing the underlying causes of these predictions.

The increasing electronisation of systems and devices makes it possible to produce and process data in real time (Bradbury & Carpizo, 2018). The application of data analysis and machine learning techniques presents an opportunity to implement **predictive**, **proactive**, **corrective** and **preventive maintenance** in a more effective manner (Scaife, 2024). Today, the use of advanced industrial electronic processors and controllers (Bradbury & Carpizo, 2018) (Klathae, 2019), in addition these, the Internet of Things (IoT), cloud computing, artificial intelligence, big data and sensor technologies, enables the detection and prediction of anomalies, problems, cyber attacks and usage patterns during the operation of a system (Errandonea & Arrizabalaga, 2020) (Gbadamosı, 2023) (Hosseini, 2021). The processing of data on the system's diverse characteristics and behavioural patterns, through the utilisation of feature selection and machine learning technologies, provides insight into the operational dynamics of the system. Understanding of the internal dynamics of a system, which may not be apparent from an external perspective, improves diagnostics and fault analysis, highlights potential areas for enhancing the system's reliability and provides insights for designing modifications to improve its performance (Turnbull, 2021) (Zhao, Li, & Zhang, 2019) (Diez-Olivan, Ser, Galar, & Sierra). In addition to this, the ability to predict system behaviour and to perform proactive maintenance activities depends on the development of predictive tools based on historical data, statistical inference methods, and engineering approaches (Nacchia, Fruggiero, Lambiase, & Bruton, 2021).

The effectiveness of machine learning models is significantly influenced by the quality of the data, the efficiency of the feature selection process, the selection of models that align with the characteristics of the problem, and the systematic implementation of the machine learning life cycle. In industrial studies, problems of access to quality data are frequently encountered (Saha, Mukherjee, Ankit, & Jadhav, 2024) (Cinieri, 2019). Missing information, the failure to create time series, a lack of maintenance histories, structural differences in data sets and data transformation losses may be considered under this heading. Additionally, the considerable data volume and the challenges associated with accessing domain expertise on the design and internal dynamics of the system, which is necessary to cope with the complexity of the system and to interpret the model results, should also be considered as potential risks. The lack of access to domain expertise restricts the interpretation of the output of data analysis, the ability to gain a comprehensive understanding of the relevant factors, the formulation of hypotheses and the generation of practical insights regarding feature selection (Victor, Schmeißer, Leitte, & Gramsch, 2024) (Srimrudhula, 2024) (Martins, 2023) (Lede, 2024) (Analytics, 2024).

In this paper, as part of a TUBITAK research project to develop a predictive maintenance (PdM) system for electric buses, the predictive performance of certain alarms in vehicles using CAN Bus data is investigated with machine learning techniques. In the study, CAN Bus data obtained from five different vehicles and transferred to the machine learning process at different times over a two-year period during the project was processed. In order to address the risks caused by this situation and to obtain expert opinion more effectively, a graph-based feature selection tool was developed, statistical relationships between variables were visualised, and the machine learning life cycle was applied in a disciplined manner, allowing iterative and cross-validation (Victor, Schmeißer, Leitte, & Gramsch, 2024). In this way, it is aimed to examine the effects of differences in the data sets of buses on modelling performance, and to provide insight into the internal dynamics of the system through the outputs of the graph-based feature selection tool and the explainability outcomes of the models. Filter, Wrapper, Embedded feature selection approaches, which are frequently used in the literature, fail to capture multiple component interactions compared to Hybrid approaches and are model-dependent approaches that can be applied iteratively in a long time (Das, Goswami, Chakrabarti, & Chakraborty, 2017) (Cheng, ve diğerleri, 2023) (Schroeder, Styp-Rekowski, Schmidt, Acker, & Kao, 2019) (Zhang, 2012). The graph-based feature selection tool created within the scope of this study aims to capture the relationships of components at multiple levels and make them observable. In this way, it is tried to reduce the complexity and model dependency of the feature selection phase and to increase its interpretability.

## 1.1 Dataset

In this study, real-time CAN bus data collected from five different electric buses between 29 September 2021 and 23 June 2023 was used. The data was collected at 1-second intervals and stored in a tabular format. Each dataset consisted of various operational parameters, performance metrics and diagnostic signals recorded as time series. The CAN bus data covered a wide range of variables, including vehicle speed, battery temperature, state of charge (SOC), motor torque, brake air pressure and other critical indicators required to predict system faults and alarms.

**Summary of the Raw Datasets**

The datasets collected from each bus varied in terms of the number of files, total size, and number of unique variables. Table 1 provides a summary of the collected datasets. For instance, the **reference dataset (3F2551C1)** contained 80.2 GB of data, comprised of 902 unique variables, and covered a vehicle operational period from September 30, 2021, to May 26, 2023. In contrast, other datasets like **EF818714** and **920CB615** varied in size and the number of variables, but similarly contained CAN Bus data collected during real-time operations. The reference dataset was selected based on expert opinion as it contained a higher frequency of target alarms and more unique variables, making it ideal for model development.

Table 1. Summary of collected dataset

| Dataset Name | First Submission Date | Dataset Size | Number of Variables | Average Frequency of Target Variables | Vehicle Operation Period |
|---|---|---|---|---|---|
| 3F2551C1 (Reference) | June 26, 2023 | 80.2 GB | 902 | 974.5 | Sept 30, 2021 - May 26, 2023 |
| 5E3DA3A9 | July 10, 2023 | 79.4 GB | 872 | 123.2 | Sept 29, 2021 - June 23, 2023 |
| 6BF5FA70 | July 17, 2023 | 26.3 GB | 875 | 30.2 | Sept 30, 2021 - Nov 10, 2022 |
| 920CB615 | August 17, 2023 | 24.8 GB | 838 | 70 | Oct 1, 2021 - June 20, 2023 |
| EF818714 | August 25, 2023 | 80.2 GB | 893 | 970 | Dec 13, 2021 - June 20, 2023 |

**Data Preprocessing and Transformation**

Once the raw data had been collected, pre-processing techniques were applied to ensure data quality and consistency. The dataset timestamps, which were initially in different formats, were converted to a standardised format using the isoparse function to comply with the ISO 8601 format. In addition, missing values were handled, and outlier records were carefully examined to improve the reliability of the dataset. These pre-processing steps were crucial in preparing the data for the subsequent analysis and model training.

**Target Variables and Alarms**

The key focus of the dataset was on predicting specific target alarms, as shown in Table 2. The alarms included critical diagnostic and general warnings such as the **ABS/EBS Amber Warning Signal (AWS)**, **Dashboard System Warning (DSW)**, **Dashboard System Fault (DSF)**, and battery management-related alarms like **PDU Fault** and **BMS_SOC**. Each of these alarms was tied to the vehicle's operational health and performance, and their early prediction was the primary goal of the machine learning models.

Table 2. List of Target Variables (Alarms)

| Variable Name | Variable Type | Variable Description |
|---|---|---|
| CAN1.EBC1.ABS_EBSAmberWarningSignal (AWS) | Diagnostic Alarm | Defines a specific condition or warning related to the vehicle's electronic braking system and safety warning systems. |
| CAN1.VMU1.Dsh_System_Warning (DSW) | General Alarm | Represents a signal indicating a system warning on the vehicle's dashboard, informing the driver of potential issues with various vehicle systems. |
| CAN1.VMU1.Dsh_System_Fault (DSF) | General Alarm | Represents a signal indicating a system fault on the vehicle's dashboard, informing the driver of a system failure. |
| CAN1.BMS_Status_Message.PDU_FAULT (PDU) | Diagnostic Alarm | Represents a signal indicating an error in the Battery Management System (BMS), warning of a serious fault or abnormality in the battery system. |
| CAN1.Battery_Status.BMS_SOC (BMS_SOC) | Diagnostic Alarm | Represents the state of charge (SOC) of the battery in the Battery Management System (BMS), indicating the battery's current energy capacity. |

## 2 Methodology

In the study, several machine learning and data processing methods were used to develop a predictive maintenance system for electric buses using CAN bus data. The raw data, collected from several buses over a period of two years, required comprehensive pre-processing techniques, including data normalisation and anomaly detection. To improve feature selection and to optimise model performance, a graph-based feature selection method was developed using statistical methods such as Pearson correlation, Cramér's V and ANOVA F-test. The method also used heuristic and optimization-based community detection algorithms (InfoMap, Leiden, Louvain, Fast Greedy) to identify significant subsets of features and reduce data dimensionality. Machine learning models, including Support Vector Machine (SVM), Random Forest (RF) and eXtreme Gradient Boosting (XGBoost), were trained and optimised using hyper-parameter tuning methods such as Grid Search and Random Search. Techniques such as SMOTEEN and binary search-based down-sampling were implemented to handle unbalanced datasets, while explainability tool, LIME, was used to interpret model predictions and identify key features.

### 2.1 Raw Data Analysis

A comprehensive data analysis was performed on the reference dataset (3F2551C1), which provides information on an electric bus designed for urban use. The purpose of this analysis was to understand the general characteristics of the data, identify potential problems and extract key insights. The first stage of the data analysis process was to examine the statistical properties of the dataset. For each of the 902 attributes in the raw data, the data type, minimum and maximum values, distribution, mean and standard deviation were calculated to understand the central tendencies and distributions of the variables. Variables' left or right skewness were identified by comparing the mean and median values. In addition, variables showing a wide distribution around the mean with a high standard deviation were identified. The variables were divided into two groups as "categorical" and "continuous". This classification was based on the data types and the number of unique values for each variable. Any variable with an integer or string data type and less than 20 unique values was classified as a categorical variable.

**Figure 1.** Machine Learning lifecycle and techniques employed in the research

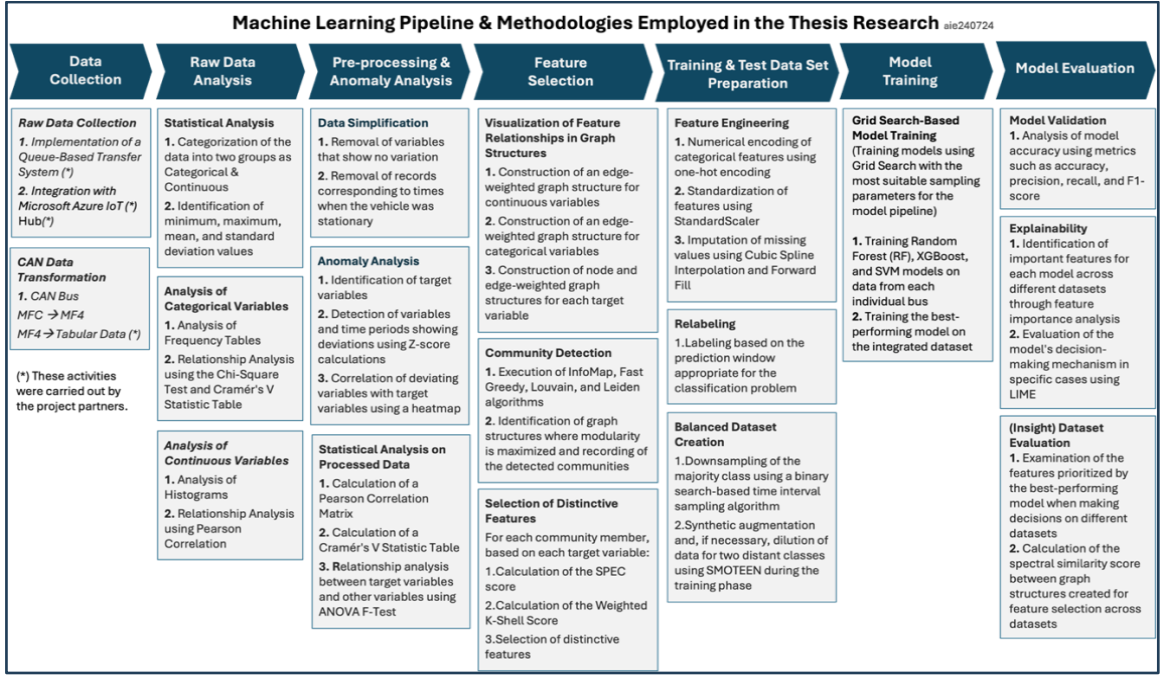

The distribution and frequencies of categorical variables were analysed using frequency tables. Cramér's V statistic, which measures the strength of association based on chi-square values, was used to examine relationships between categorical variables shown in Equation 1. A relationship matrix was created to visualise the associations, highlighting stronger relationships between certain variables. This matrix also allowed analysis of changes in variable relationships over the two years of vehicle operation, including both stationary and moving phases.

$$V = \sqrt{\left(\frac{(x^2/n)}{(min(k-1, r-1))}\right)} \quad (1)$$

Where; $x^2$ : Chi-Square Value; $n$ : Total Sample Size; $k$ : is the number of categories (columns); $r$ : is number of rows. This formula is used to measure the strength of association between two categorical variables.

In the case of continuous variables, histograms of the 48 variables with the highest standard deviation were plotted to examine the patterns of distribution. Skewed distributions and differences between mean and median were identified. Pearson correlation was used to measure relationships between continuous variables, with a matrix visualising the direction and strength of associations. The matrix also provided a method of examining how these relationships evolved over the same two-year period. This analysis provided insight into the behaviour of the variables and aided in the development of predictive maintenance models.

## 2.2 Data Preprocessing and Anomaly Analysis

The electrical CAN bus data set is large and complex, requiring extensive pre-processing for machine learning models. In this phase, the raw data was simplified and transformed to create a representative set of variables. The aim was to improve data quality, reduce the complexity of feature selection and improve model performance.

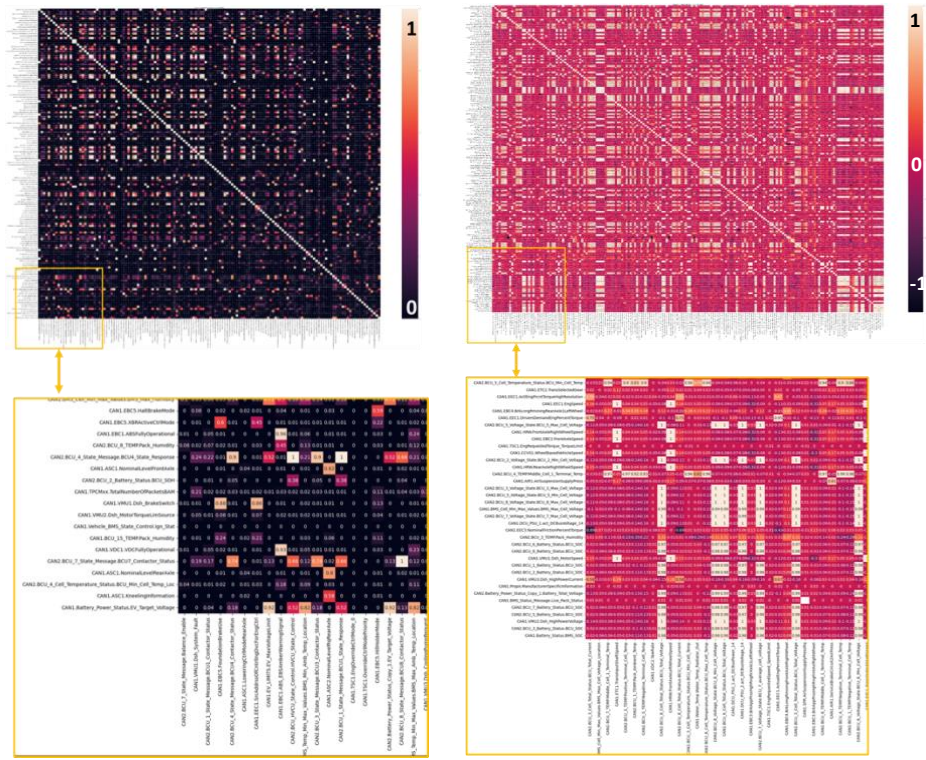

**Figure 2.** Association matrices of categorical and contionus variables

**Data Preprocessing**

Continuous variables that showed no variation and categorical variables that contained a single value across all datasets were identified through statistical analysis and removed from the dataset. In addition, variables related to CAN bus communication protocols, such as message size and checksum codes, were excluded as they were considered insignificant for improving model performance based on expert feedback. The number of records was reduced by filtering out data from periods when the vehicle was stationary, as these records did not contribute to feature selection or model training in a meaningful way and could potentially reduce model performance. Variables such as vehicle speed and motor speed were used to identify stationary periods, and the corresponding records were eliminated to improve data quality and the efficiency of the following analysis.

**Anomaly Analysis**

This phase focused on identifying unusual patterns or anomalies in the CAN bus data set that could indicate potential problems affecting the performance, safety or lifetime of the electric bus. This phase was critical because the complex interactions between the vehicle's systems and components make it difficult to identify which parts were affected by unexpected events. In that step, relationships between critical system alarms and sensor data, battery management system data and other vehicle components examine to identify anomalies that could have affected performance.

The first step in the anomaly analysis was to identify the periods when target alarms related to the braking, transmission and battery systems were active. These alarms, which were considered critical due to their direct impact on key vehicle components, were monitored for periods of five seconds or more. The corresponding time intervals were recorded for further analysis. The second step was to identify the times when other system and battery variables deviated significantly from normal behaviour. In that analysis the variables were divided into two groups: system variables and battery variables. Using GPU-based libraries (cuDF and cuPy), Z-scores were calculated for each variable to quantify how much the variable deviated from its expected behaviour. A z-score threshold of 3 (absolute value) was used to identify significant anomalies, as values greater than 3 indicate that a variable deviates significantly from its normal range.

During this process, the relationships between active alarms and variable anomalies were closely examined. For example, periods when the **Dsh_System_Warning** and **Dsh_System_Fault** alarms were triggered showed strong correlations with anomalies in both system and battery variables. These anomalies were often associated with brake system components (such as brake pad thickness) and battery related variables, indicating that the alarms were triggered by significant deviations in these components. Furthermore, the **PDU_Fault** alarm analysis revealed a strong relationship between anomalies in vehicle motion and braking system variables (**AirSuspensionSupplyPress**, **SpeedLimit, TorqueLimit**) and battery monitoring variables. The alignment of the anomalies in both the braking and battery systems suggested that **regenerative braking** was actively engaged during these periods, a conclusion confirmed by the anomaly analysis. The **BMS_SOC** alarm, which monitors the state of charge (SOC) of the battery, also showed a strong correlation with both braking system variables and battery-charging monitoring variables during certain periods. This alarm had a direct impact on the vehicle's motion and braking systems, showing how deviations in SOC levels were linked to anomalies in these critical systems.

Finally, statistical analyses were performed on the **refined** dataset. Pearson correlation coefficients were recalculated for continuous variables to assess their relationships, while Cramér's V statistic was used to assess associations between categorical variables. In addition, ANOVA F-tests were used to examine the relationships between target alarms and both continuous and categorical variables.

### 2.3 Graph-Based Future Selection

Graph-based feature selection was used to reduce the dataset size and improve model performance. In this approach, statistical relationships between variables were visualised in a graph structure and community detection algorithms (Leiden, Louvain, InfoMap, Fast Greedy) were applied to identify relevant feature subsets. Then, based on graph theory principles, discriminative features were identified by calculating metrics such as SPEC score and K-shell decomposition. This method effectively reduced the dimensionality of the dataset and aimed to improve the performance of the predictive models.

In this study, a feature selection algorithm combining statistical filtering methods with graph-based principles was developed to identify distinctive features (attributes) from the simplified CAN bus data. The developed algorithm constructed weighted graphs for each target alarm, where node and edge weights were determined using statistical measures. By applying heuristic, greedy and optimisation-based community detection algorithms, an optimal graph structure was formed and the Relevant Feature Subsets (RFS) represented by the detected communities were identified. Finally, the nodes within each RFS were scored using the Spectral Feature Selection (SPEC) score and a modified K-shell score - adapted from the methodology in (Cheng, ve diğerleri, 2023) - which considers node weights. The most informative and discriminative features were then selected based on these metrics.

**Figure 3.** Graph structure of categorical variables

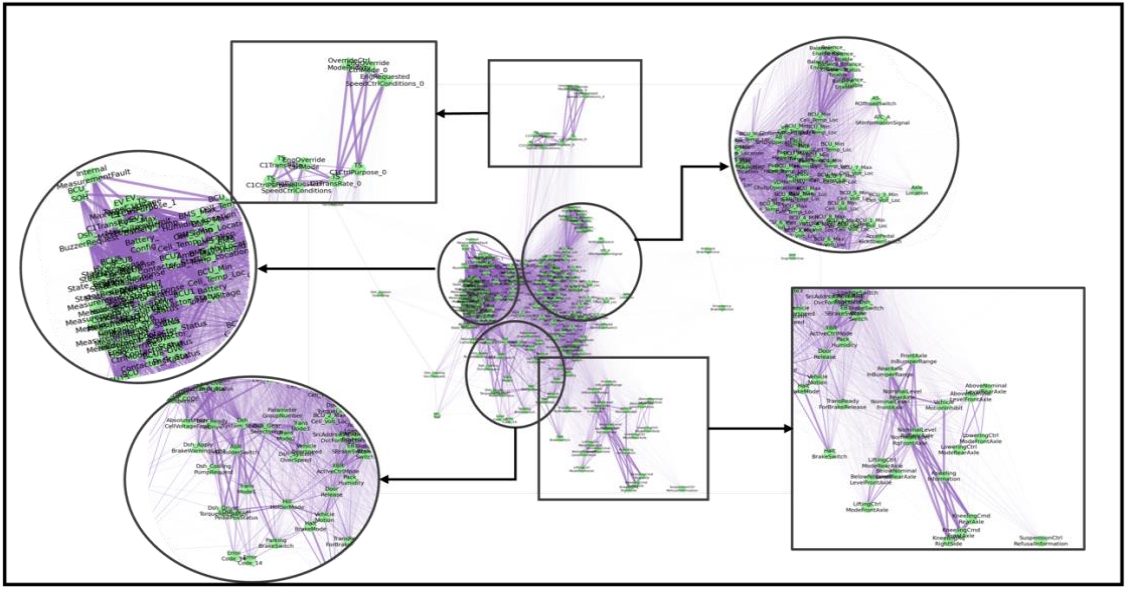

**Visualization of Statistical Relationships Between Variables**

In this phase, graph structures were created to visualize the relationships between categorical and continuous variables based on statistical analyses performed on the simplified dataset (Pearson Correlation Matrix, Cramer's V Statistic Table, and ANOVA F-Score calculations). In these graph structures, each node represents a variable, and the edges between nodes depict the strength of the statistical relationships between the variables. The relationships between continuous variables were determined using Pearson correlation coefficients, while those between categorical variables were determined using Cramer's V coefficients. As can be seen in Figure 3 and Figure 4, the thickness of the edges was adjusted in proportion to the strength of the relationship which makes stronger connections more prominent. Similarly, the distance between nodes was inversely proportional to the strength of the relationships, so variables with stronger correlations were placed closer together in the graph. In addition, the relationship between each target alarm and other variables was calculated using the ANOVA F-test, with the size of the nodes representing the strength of the relationship between categorical or continuous variables and the target alarm. Larger nodes indicated stronger relationships with the alarms, while smaller nodes represented weaker connections. This visualization provided a holistic map of the statistical relationships between the variables and the alarms, providing a clearer understanding of the complex interconnections. Particularly, dense areas of the graph and thicker edges highlighted strong relationships and potential interactions. This approach simplified the identification and interpretation of relevant feature subsets by offering a comprehensive view of the relationships across the dataset.

**Figure 4.** Visualisation of continuous variable & targeted alarm relationships

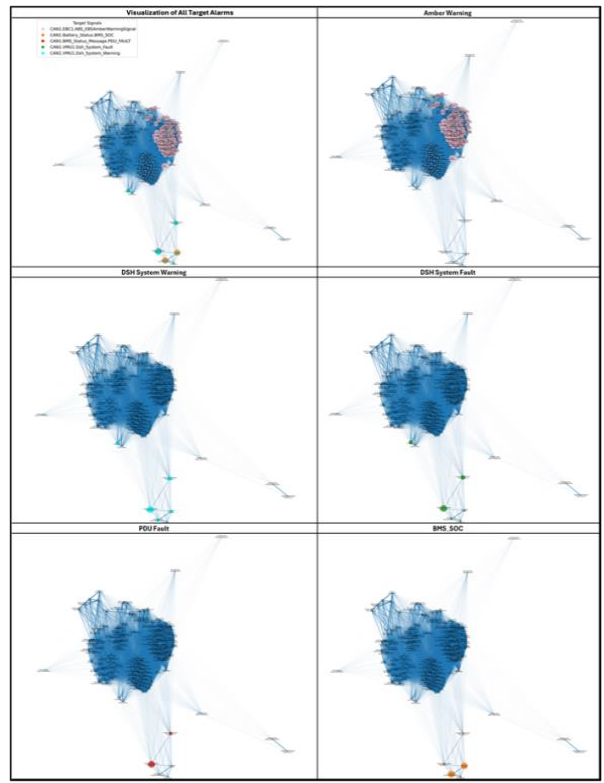

**Community Detection on Graph Structures**

Community detection algorithms were applied to the weighted graph structures to identify relevant feature subsets. The algorithms used included **InfoMap** (optimization-based), **Fast Greedy** (greedy approach), and **Louvain** and **Leiden** (heuristic approaches). The quality of the resulting communities was evaluated using the **modularity** metric. Unlike traditional methods that assign a fixed threshold to remove edges below a certain value, a more dynamic and optimization-focused approach was adopted to enhance the applicability of the method as it can be seen in Figure 5 and Algorithm 1.

**Algorithm1 & Figure 5.** Community detection with dynamic threshold algorithm and workflow

```
Algorithm 1 Community Detection with Dynamic Thresholds
 1: function COMMDETECTDYNTHRESHOLDS(Graph G)
 2:     best_modularity ← −1
 3:     best_community_structure ← None
 4:     best_threshold ← None
 5:     best_algorithm ← None
 6:     thresholds ← [0.1, 0.15, 0.2, ..., 0.95]
 7:     algorithms ← ['Leiden', 'InfoMap', 'FastGreedy', 'Louvain']
 8:     for each threshold ∈ thresholds do
 9:         G_thresholded ← REMOVEEDGES(G, threshold)
10:         for each algo ∈ algorithms do
11:             communities ← COMMUNITYDETECTION(G_thresholded, algo)
12:             modularity ← MODULARITY(G_thresholded, communities)
13:             if modularity > best_modularity then
14:                 best_modularity ← modularity
15:                 best_community_structure ← communities
16:                 best_threshold ← threshold
17:                 best_algorithm ← algo
18:             end if
19:         end for
20:     end for
21:     return best_community_structure, best_threshold, best_algorithm
22: end function
23:
24: function REMOVEEDGES(Graph G, float threshold)
25:     new_graph ← Copy of G
26:     for each (u, v, weight) ∈ G.edges(data='weight') do
27:         if weight < threshold then
28:             new_graph.remove_edge(u, v)
29:         end if
30:     end for
31:     return new_graph
32: end function
33:
34: function COMMUNITYDETECTION(Graph G, string algorithm)
35:     if algorithm == 'Leiden' then
36:         return APPLYLEIDEN(G)
37:     else if algorithm == 'InfoMap' then
38:         return APPLYINFOMAP(G)
39:     else if algorithm == 'FastGreedy' then
40:         return APPLYFASTGREEDY(G)
41:     else if algorithm == 'Louvain' then
42:         return APPLYLOUVAIN(G)
43:     else
44:         raise ValueError("Unknown algorithm")
45:     end if
46: end function
47:
48: function ModularityGraph G, communities
49:     Q ← 0
50:     m ← total weight of all edges in G
51:     for each community in communities do
52:         L_c ← total weight of edges within community
53:         D_c ← total sum of weights of edges attached to nodes in community
54:         $Q \leftarrow Q + \left(\frac{L_c}{m} - \left(\frac{D_c}{2m}\right)^2\right)$
55:     end for
56:     return Q
57: end function
```

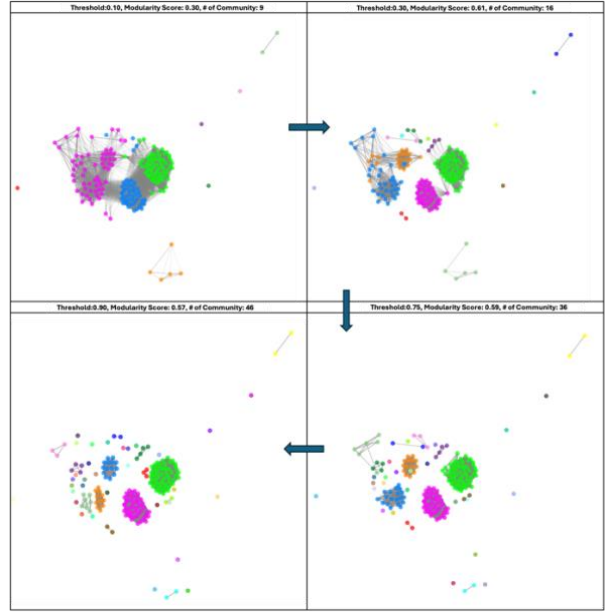

In this approach, thresholds ranging from 0.1 to 0.95 were applied incrementally in steps of 0.05. For each threshold, edges below the corresponding value were removed from the graph and the community detection algorithms (Leiden, InfoMap, Fast Greedy and Louvain) were executed. The resulting community structures were evaluated based on their modularity scores. The combination of algorithms and thresholds that gave the highest modularity score was selected, which helped to identify the optimal communities on the graphs. This approach contributed to a deeper understanding of the relationships between variables and highlighted relevant subgroups (communities), which were stored for later use in the feature selection process. As can be seen in Table 3, for the graph of continuous variables, a threshold of 0.30 and the application of the **Leiden** algorithm produced the best modularity score of 0.6133, resulting in a graph with 16 communities. On the other hand, for the graph of categorical variables, a threshold of 0.95 and the use of either the **Louvain** or **InfoMap** algorithms yielded the highest modularity score of 0.6222, producing a graph with 102 communities. This contrast demonstrated that categorical variables exhibited a more heterogeneous structure compared to continuous variables, and that both types of variables followed distinct relational patterns.

Table 3. Performance of community detection algorithms

| Algorithm Performance on Graph of Continous Variables | | | |
|---|---|---|---|
| Community Detection Algorithm | Modularity Score | Threshold | Number of Communities |
| InfoMap | 0.5961 | 0.40 | 22 |
| Fast Greedy | 0.5999 | 0.35 | 16 |
| Louvein | 0.6106 | 0.30 | 16 |
| **Leiden** | **0.6133** | **0.30** | **16** |
| Algorithm Performance on Graph of Categorical Variables | | | |
| Community Detection Algorithm | Modularity Score | Threshold | Number of Communities |
| **InfoMap** | **0.6222** | **0.95** | **102** |
| Fast Greedy | 0.6210 | 0.95 | 101 |
| **Louvein** | **0.6222** | **0.95** | **102** |
| Leiden | 0.6216 | 0.95 | 102 |

**Selection of Distinctive Features**

In this stage, distinctive features were identified by calculating the **Spectral Feature Selection (SPEC) score** and a modified **K-Shell score** for variables within the communities identified from the graph structures. For continuous variables, a threshold of 0.30 was applied along with the **Leiden** algorithm, while for categorical variables, a threshold of 0.95 and the **InfoMap** algorithm were used. The relationships between the variables within the communities and the target alarms were considered to determine which features were most distinctive.

*Calculation of the SPEC Score*

The **Spectral Feature (SPEC) score** was used to assess the importance of features within the communities and their relationships with the target alarms. This score evaluates the similarity of the features to others within the same community, while also considering their relationship with the target alarms.

First, the **Laplacian matrix** of the subgraph structure, representing the nodes and edges within each community, was constructed, as defined by Equation (2). The Laplacian matrix captures the structural properties and relationships between the nodes in the graph.

$$L = D - A \tag{2}$$

*Where D: A diagonal matrix that contains the number of neighbors (degrees) of each node in the graph, A: The adjacency matrix of the graph.*

$$SPEC(x_i) = \sum_{\{j=1\}}^{k} \lambda_j \left(u_j^T x_i\right)^2 \times F_i \tag{3}$$

*Where, $\lambda\_j$: The j-th eigenvalue of the Laplacian matrix, $u\_j$: The j-th eigenvector of the Laplacian matrix, $x\_i$: The i-th feature vector, $F\_i$: The ANOVA F-score of the i-th feature.*

Next, the **eigenvalues** and **eigenvectors** of the Laplacian matrix were computed to represent the relationships between the nodes. The ANOVA F-scores, previously calculated for each feature with respect to each target variable, were then incorporated into the eigenvectors. Specifically, the F-scores for each feature and target variable were vectorized and multiplied by the corresponding eigenvectors of the Laplacian matrix, weighting the eigenvectors by the F-scores. This operation resulted in new eigenvectors that reflected both the structural position of the features within the graph and their relationship with the target variables.

Finally, the sum of the squares of these newly weighted eigenvectors was computed to calculate the SPEC score for each feature within the community. This score indicated which features had strong relationships with others within the same community and maintained significant connections with the target variables. The SPEC score was calculated as shown in Equation 3, allowing the identification of features that were both structurally and statistically relevant within the graph.

*Calculation of the Modified K-Shell Score*

In this stage, the ANOVA F-score values, which represent the relationship between features and alarms, were used as node weights. Drawing on the method described in (Cheng, ve diğerleri, 2023), the **modified K-Shell score** for each feature within the communities was calculated for each alarm, as shown in Equation 4. The modified K-Shell score accounts for both the weight of the feature itself and the weights of its neighbors, as well as the strength of the connections between them. This score evaluates the importance of each feature in the community by considering both its relationship with the alarms and its centrality within the community structure.

$$k'_{\{\eta\}} = \left[ \sqrt{k_{\{\eta\}} W_{\{V_{\{\eta\}}\}} \Sigma_{\{k_{\{\eta\}}\}} W_{\{E_{\{\eta,\xi\}}\}}} \right] \quad (4)$$

*Where;* $k_{\{\eta\}}$: *This represents the degree of the node* **η**, *which is the number of edges connected to the node.* $W_{\{V_{\{\eta\}}\}}$: *The weight of the node* **η**, $W_{\{E_{\{\eta,\xi\}}\}}$: *The sum of the weights of the edges between node* **η** *and the other connected nodes*

*Identification of Distinctive Features from Communities*

At this stage, the features within each community that had the highest SPEC and K-Shell scores for the target alarms were selected, resulting in two potential sets of distinctive features. It was observed that the set obtained by selecting the features with the highest K-Shell scores contained fewer features compared to the set selected based on the maximum SPEC scores. This difference was attributed to the lower performance of the K-Shell decomposition method in communities with fewer features. On the other hand, when trying to identify the most distinctive feature from those with the maximum SPEC score in a community, the K-Shell score was used to select the feature that had the most relationships with other community members and held the most central position. In this way, a single, unified set of distinctive features was created by prioritizing features that were both statistically significant and structurally central within their communities.

**Table 4.** Feature Sets Created According to The SPEC Score & K-Shell Score.

| Community Name | Selected Features According to SPAC Values | Community Name | Selected Features According to K-Shell Values |
|---|---|---|---|
| Community 1 | CAN1.Water_Temp.Water_Temp_Radiator_Out<br>CAN1.VMU1.Dsh_ThermalIndicator<br>CAN1.VMU2.Dsh_AuxVoltage | Community 1 | CAN1.Water_Temp.Water_Temp_Radiator_Out<br>CAN2.BCU_7_TEMP.Positive_Terminal_Cell_Temp<br>CAN1.VMU2.Dsh_AuxVoltage<br>CAN2.BCU_4_TEMP.Pack_Ambient_Temp |
| Community 2 | CAN1.BMS_Cell_Min_Max_Values.BMS_Max_Cell_Voltage | Community 2 | CAN1.BMS_Cell_Min_Max_Values.BMS_Max_Cell_Voltage |
| Community 3 | CAN1.VMU3.Dsh_HighPowerCurrent<br>CAN1.ASC4.BellowPressRearAxleRight | Community 3 | CAN1.ASC4.BellowPressRearAxleLeft |
| Community 4 | CAN1.DCU_PSU_2.act_DCBusPower_14<br>CAN1.EBC1.BrakePedalPos | Community 4 | CAN1.DCU_PSU_2.act_DCBusPower_14 |
| Community 5 | CAN1.TSC1.EngRequestedTorqueHighResolution_0<br>CAN1.TSC1.EngRequestedSpeed_SpeedLimit_1<br>CAN1.TSC1.EngRequestedTorque_TorqueLimit_2 | Community 5 | CAN1.TSC1.EngRequestedSpeed_SpeedLimit_1<br>CAN1.TSC1.EngRequestedTorqueHighResolution_0 |
| Community 6 | CAN1.EBC4.BrkLnngRmningFrontAxleRightWheel<br>CAN1.EBC4.BrkLnngRmningFrontAxleLeftWheel<br>CAN1.VD.TotalVehicleDistance | Community 6 | CAN1.EBC4.BrkLnngRmningFrontAxleRightWheel |
| Community 7 | CAN1.AIR1.AirSuspensionSupplyPress<br>CAN1.AIR1.ServiceBrakeCircuit2AirPress | Community 7 | (-) |
| Community 8 | CAN1.DCU_PSU_1.act_DCBusPower_14 | Community 8 | (-) |
| Community 9 | (-) | Community 9 | (-) |
| Community 10 | CAN1.ETC2.TransCurrentGear | Community 10 | (-) |
| Community 11 | CAN1.BMS_Cell_Min_Max_Values.BMS_Max_Cell_Voltage_Location | Community 11 | (-) |
| Community 12 | (-) | Community 12 | (-) |
| Community 13 | (-) | Community 13 | (-) |
| Community 14 | CAN1.BMS_Cell_Min_Max_Values.BMS_Min_Cell_Voltage_Location | Community 14 | (-) |
| Community 15 | (-) | Community 15 | (-) |
| Community 16 | (-) | Community 16 | (-) |

This final set of distinctive features was reviewed and approved by domain experts. Based on their feedback, eight features related to specific battery packs were removed from the set, as these features could vary depending on the vehicle structure. In conclusion, after all the steps of analysis and graph-based feature selection, a total of 52 distinctive features were identified from the initial raw dataset, which contained 902 variables, to be used for alarm prediction.

## 2.4 Training and Test Dataset Preparation

After graph-based feature selection, the identified, distinctive features were extracted from the simplified dataset and a comprehensive data processing and transformation workflow was performed. The objective of this workflow was to create a balanced and reliable dataset for model training, ensuring that machine learning models (RF, SVM, XGBoost) could effectively and accurately predict the target variables.
First, pre-processing steps were applied to ensure consistency and comparability of the input features. This included imputing missing values, converting categorical variables using one-hot encoding, and standardising the data to achieve uniform scales. For continuous variables, missing values were imputed using cubic spline interpolation, while categorical variables were imputed using a forward fill technique to avoid performance degradation due to missing data.

In the second phase, the dataset was restructured for compatibility with the predictive models. Label shifting was applied to predict the target alarm activations. To address the class imbalance caused by the low frequency of alarm activations, a **binary search-based down-sampling algorithm** was developed, as shown in Algorithm 2 &3. This method selectively down-sampled records where alarms were inactive, reducing the dominance of non-alarm data while preserving the temporal structure of the dataset. Tehn, SMOTEEN (Synthetic Minority Over-sampling Technique for Nominal and Continuous) was used to balance the data, and stratified sampling ensured that the class distributions were preserved in both the training and test datasets.

**Algorithm 2 & 3.** Dynamic Threshold-Based Binary Search Down-Sampling Algorithm

```
Algorithm 2 Time Interval Undersampling
Require: Data data, Time Column time_column, Interval interval, Number
    of Samples num_sample
Ensure: Undersampled Data X, Target Variable y
 1: Set df ← data
 2: Set df[target_name] ← y
 3: Set df ← df.set_index(time_column)
 4: Initialize undersampled_data ← []
 5: for all groups g in df.resample(interval) do
 6:     if length of g > num_sample then
 7:         Append g.sample(num_sample) to undersampled_data
 8:     else
 9:         Set updated_sample ← length of g − num_sample
10:         Append g.sample(updated_sample) to undersampled_data
11:     end if
12: end for
13: Set undersampled_df ← pd.concat(undersampled_data)
14: Set undersampled_X ← undersampled_df.drop([target_name], axis = 1)
15: Set undersampled_y ← undersampled_df[target_name]
16: return undersampled_X, undersampled_y
```

```
Algorithm 3 Find Optimal Interval
Require: Data data, Target Size target_size, Minimum Interval min_interval,
    Maximum Interval max_interval
Ensure: Best Interval best_interval
 1: Set left ← min_interval
 2: Set right ← max_interval
 3: Set best_interval ← max_interval
 4: while left ≤ right do
 5:     Set mid ← (left+right)/2
 6:     Set undersampled_data, _ ← Time Interval Undersampling(data, 'times-
        tamps', midT)
 7:     if length of undersampled_data ≥ target_size then
 8:         Set best_interval ← mid
 9:         Set left ← mid + 1
10:     else
11:         Set right ← mid − 1
12:     end if
13: end while
14: return best_interval
```

To address the class imbalance, particularly the lack of representation of alarm activations, a binary search based down-sampling algorithm was used. This algorithm dynamically identified an optimal time interval for down-sampling non-alarm records in order to balance the number of active and inactive alarm records and the time series structure of the dataset was preserved to ensure temporal consistency. The algorithm first divided the dataset into time intervals, grouping records according to the activation status of alarms. From these groups, random samples of non-alarm records were taken and iteratively reduced to approximate the number of alarm records. To determine the optimal time interval for the down-sampling, the algorithm performed a binary search over a range of time intervals defined by minimum and maximum values. The target data size was set to match the number of alarm-activated records. By running the developed binary search algorithm, the optimal time window was identified that also allowed for effective down-sampling without compromising the integrity of the dataset.

**Creation of Training and Test Sets**

After downsampling, the balanced dataset was divided into training and test sets, with 70% assigned to training and 30% to testing. Stratified sampling was used to maintain proportional class distributions in both datasets, ensuring that both training and test sets contained similar distributions of alarm activations. This step was critical to increase the generalisation of the models and ensure consistent performance across the datasets.

## 2.5   Model Training

n this study, Support Vector Machines (SVM), Random Forest (RF) and Extreme Gradient Boosting (XGBoost) models were trained on both the training and test datasets to predict alarm activation. The SMOTEEN (Synthetic Minority Oversampling Technique-Edited Nearest Neighbours) method was used to address class imbalance. This technique generated synthetic examples for the minority class while removing misclassified instances to improve the overall balance and quality of the dataset.

A pipeline architecture was used for model training, integrating both SMOTEEN and the models to facilitate hyper-parameter optimisation via GridSearchCV. This allowed the parameters of the models and the parameters of the SMOTEEN technique (k_neighbors and enn_n_neighbors) to be tuned simultaneously.

This approach ensured that the models performed optimally by finding the best combinations of parameters.

The training process was divided into two phases. First, the models were trained on a reference dataset from the 3F2551C1 dataset, with the goal of predicting alarm activations 20 minutes in advance. After achieving high performance on the reference dataset Table 5, the models were further evaluated using datasets from other buses to test generalization. However, the results (Table 6) showed that models trained on the reference dataset did not generalize well to the other buses, primarily due to differences in the data characteristics between buses. To mitigate this, a generalized model was developed by combining training sets from both the reference bus and other buses, which significantly improved performance across different bus datasets (Table 7).

Table 5. Model Performances on Reference Data Set

| **CAN1.EBC1.ABS_EBSAmberWarningSignal** | | | | |
|---|---|---|---|---|
| **Model** | **ACC** | **PRE** | **REC** | **F1** |
| SVM | 0.99 | 0.97 | 0.98 | 0.99 |
| RF | 0.98 | 0.96 | 0.97 | 0.98 |
| XGBoost | 0.99 | 0.98 | 0.99 | 0.98 |
| **CAN1.VMU1.Dsh System Fault** | | | | |
| **Model** | **ACC** | **PRE** | **REC** | **F1** |
| SVM | 0.98 | 0.99 | 0.98 | 0.99 |
| RF | 0.99 | 0.99 | 1.00 | 1.00 |
| XGBoost | 1.00 | 0.99 | 1.00 | 0.99 |
| **CAN1.VMU1.Dsh System Warning** | | | | |
| **Model** | **ACC** | **PRE** | **REC** | **F1** |
| SVM | 0.98 | 0.99 | 0.98 | 0.99 |
| RF | 0.99 | 0.99 | 1.00 | 1.00 |
| XGBoost | 1.00 | 0.99 | 1.00 | 0.99 |
| **CAN1.BMS_Status_Message.PDU_FAULT** | | | | |
| **Model** | **ACC** | **PRE** | **REC** | **F1** |
| SVM | 0.99 | 1.00 | 1.00 | 1.00 |
| RF | 0.99 | 1.00 | 1.00 | 1.00 |
| XGBoost | 0.99 | 1.00 | 1.00 | 1.00 |
| **CAN1.BMS_Status_Message.BMS_SOC** | | | | |
| **Model** | **ACC** | **PRE** | **REC** | **F1** |
| SVM | 0.97 | 0.97 | 0.99 | 0.9 |
| RF | 1.00 | 0.99 | 1.00 | 1.00 |
| XGBoost | 1.00 | 1.00 | 1.00 | 1.00 |

Table 6. The Test Results of The Model Trained with Reference Bus Data Using the Data from Other Buses.

| | 5E3DA3A9 | | 6BF5FA70 | | 920CB615 | | EF818714 | |
|---|---|---|---|---|---|---|---|---|
| | ACC | REC | ACC | REC | ACC | REC | ACC | REC |
| CAN1.EBC1.ABS_EBSAmberWarningSignal | 0.58 | 0.08 | 0.51 | 0.00 | 0.50 | 0.00 | 0.62 | 0.01 |
| CAN1.VMU1.Dsh System Fault | -- | -- | -- | -- | -- | -- | -- | -- |
| CAN1.VMU1.Dsh System Warning | 0.56 | 0.06 | 0.52 | 0.00 | -- | -- | 0.54 | 0.05 |
| CAN1.BMS_Status_Message.PDU_FAULT | -- | -- | -- | -- | -- | -- | -- | -- |
| CAN1.BMS_Status_Message.BMS_SOC | 0.55 | 0.03 | 0.50 | 0.00 | -- | -- | 0.51 | 0.00 |

Table 7. Results of The Generalised Model Training

|  | 5E3DA3A9 | | 6BF5FA70 | | 920CB615 | | EF818714 | |
|---|---|---|---|---|---|---|---|---|
|  | ACC | REC | ACC | REC | ACC | REC | ACC | REC |
| CAN1.EBC1.ABS_EBSAmberWarningSignal | 0.98 | 0.98 | 0.99 | 1.00 | 1.00 | 1.00 | 0.96 | 0.92 |
| CAN1.VMU1.Dsh System Fault | -- | -- | -- | -- | -- | -- | -- | -- |
| CAN1.VMU1.Dsh System Warning | 0.90 | 0.87 | 0.99 | 0.98 | -- | -- | 0.95 | 0.89 |
| CAN1.BMS_Status_Message.PDU_FAULT | -- | -- | -- | -- | -- | -- | -- | -- |
| CAN1.BMS_Status_Message.BMS_SOC | 0.99 | 0.99 | 0.97 | 0.96 | -- | -- | 0.90 | 0.95 |

## 3  Findings & Explainability

During the evaluation process, differences between the datasets from various buses were observed, which were critical in understanding the models' performance issues. In **Table 8**, refined graph structures of the datasets were created by analyzing the relationships between variables and running community detection algorithms on these structures. The modularity scores for these refined graphs indicated varying degrees of related feature sets across the different buses. The differences in these graph structures revealed that the buses had distinct data patterns, which partially explained the models' lower performance when transferring knowledge from one bus to another. Further analysis, as presented in Table 9 involved calculating the **Spectral Similarity Scores** to measure the similarity between the reference bus dataset and those of other buses. The results showed that buses with more data (i.e., larger datasets) were more similar to the reference bus, while buses with less data exhibited more distinct characteristics. This helped explain why models trained on the reference bus data had difficulty generalizing to other buses with lower spectral similarity. By incorporating the findings from Table 8 and Table 9, it became evident that the **differences in data characteristics** between buses played a significant role in model performance. These insights led to the development of a **generalized model**, which combined data from multiple buses to ensure better predictive capabilities across all datasets. This generalized approach significantly improved model accuracy and interpretability by accounting for the inherent differences between the bus datasets.

Table 8. Refined Graph Structures of Buses

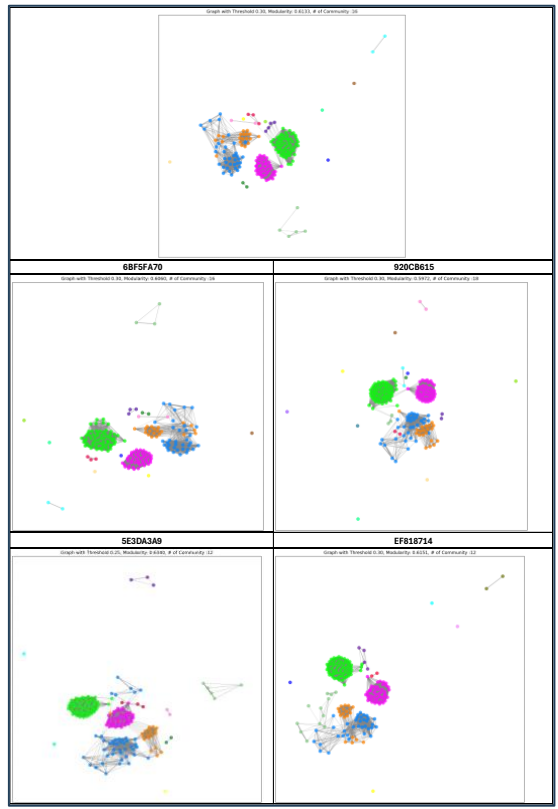

**Table 9.** The Spectral Similarity Table Showing the Similarity of Other Buses to the Reference Bus.

| Dataset Name | Spectral Similarity Score to Reference Dataset |
|---|---|
| 5E3DA3A9 | 186.66 |
| 6BF5FA70 | 253.30 |
| 920CB615 | 196.03 |
| EF818714 | 166.40 |

The explainability of the trained models was explored using **LIME** (Local Interpretable Model-Agnostic Explanations), which provides insight into the features that were most influential in the models' predictions. For each alert, LIME was applied to specific examples, highlighting the features on which the models relied most strongly. For example, as seen in Figure 6, the **Water_Temp_Radiator_Out** feature was identified as a key factor in predicting the **Amber** alert, indicating the importance of engine overheating in the model's decision-making process. Similarly, features related to battery health such as **BMS_Max_Cell_Voltage** and **BMS_Max_Humidity** were critical in predicting the **BMS_SOC** alarm, reflecting the model's focus on battery related metrics.

**Figure 6.** Explanation of Alarm Predictions Using LIME

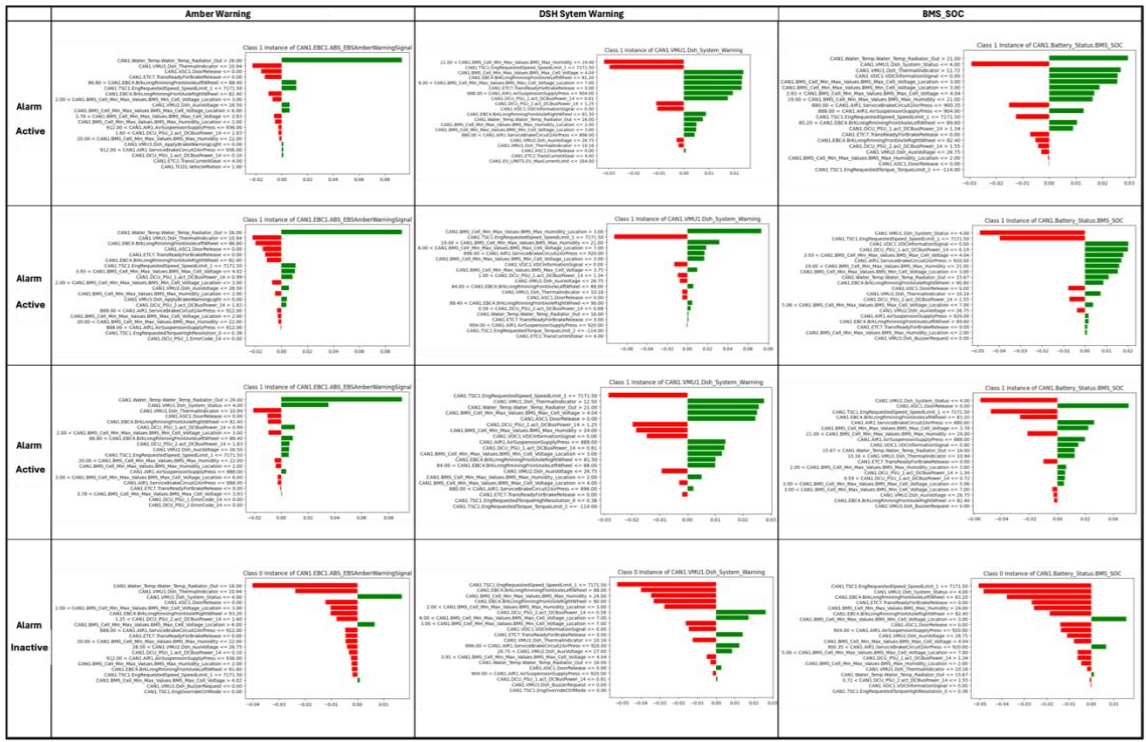

In addition, **feature importance scores** were calculated for RF and XGBoost, which verified that certain features consistently influenced model decisions across different alarms. For example, **brake pad thickness** and **battery voltage** were commonly identified as important variables, which is consistent with the findings from the anomaly analysis. These results not only validated the internal reasoning of the models, but also demonstrated consistency with domain knowledge of the bus systems.

In conclusion, the models developed for alarm prediction were highly effective when trained with balanced and representative datasets. The use of LIME and feature importance analysis improved the interpretability of the models and made their decision-making processes more transparent. The generalised model incorporating data from multiple buses showed strong predictive performance and provided solid insights into the variables most associated with alarm activations.

# 4   Conclusion

In this study, a comprehensive machine learning approach was developed to support predictive, corrective, preventive, and proactive maintenance strategies, with a primary focus on predictive maintenance. Extensive preprocessing techniques were applied to raw data packets, and hybrid graph-based feature selection tools were developed using statistical filtering methods such as Pearson correlation analysis, Cramér's V statistics, and ANOVA F-tests, along with optimization-based community detection algorithms such as InfoMap, Leiden, Louvain, and Fast Greedy. Using the developed tool, relevant feature subsets were identified, and distinctive features were selected based on their relationships with the target alarms, resulting in dimensionality reduction.

The performance of the selected features was evaluated using Support Vector Machines (SVM), Random Forest (RF), and Extreme Gradient Boosting (XGBoost) models. To address class imbalance and minimize its impact on model performance, SMOTEEN (Synthetic Minority Oversampling Technique-Edited Nearest Neighbors) and a binary search-based time interval down-sampling method were employed. Additionally, Grid Search and Random Search hyperparameter tuning methods were used to identify the optimal classifier space for each model. The performance of the developed models was evaluated based on validation metrics, ensuring robustness.

The predictions of the best-performing models were further explained by analyzing feature importance and using the LIME (Local Interpretable Model-Agnostic Explanations) tool, which provided insights into the key features driving predictions. This allowed for a deeper understanding of the system's operation and the features that triggered the alarm predictions.

Throughout the project, several real-world challenges typical of machine learning applications were managed, such as variations in field data, data quality, large data volumes, the complexity of the bus system, difficulties in interpreting relationships between variables, limited access to expert opinions, and computational constraints. To mitigate these risks, the machine learning lifecycle was carefully planned and executed in an iterative manner, allowing for cross-checks and disciplined implementation. A wide range of techniques were employed to analyze the datasets, with a strong emphasis on validating results and developing a graph-based feature selection tool to visually interpret the relationships between features.

The models achieved alarm predictions with at least 97% accuracy when trained on both bus-specific and generalized datasets, providing strong predictive capabilities 20 minutes prior to alarm activations. The performance of different models was measured using validation metrics, and the decision-making processes were explained through LIME and feature importance analysis. This approach contributed valuable insights into the system's operations, enabling effective and interpretable predictive maintenance solutions for electric buses.

# 5   Recommendations for Future Studies

This study explored graph-based feature selection, machine learning model selection, and performance evaluation using explainability methods on limited real-world data. Future research could focus on applying these methods to larger and richer datasets, which may further enhance performance and provide an opportunity to test the robustness and generalizability of the techniques across different real-world problems.

For future work in predictive maintenance using machine learning, improving the data collection and preparation processes will have a direct impact on model performance. Ensuring that vehicle data is collected comprehensively and in sufficient quantities, enhancing data quality, and enriching the dataset with non-CAN Bus data will allow models to better learn real-world conditions. Balancing the dataset and controlling model outputs will increase reliability, while maintaining traceability between system configuration differences and datasets will improve result interpretability.

The use of graph-based methods for feature selection and analysis will enable a deeper understanding of data. In larger and more complex graph-based feature representations, techniques such as hypergraphs and maximum cliques are expected to offer more effective solutions. Planning the machine learning lifecycle according to project risks and applying it in a disciplined manner is fundamental to developing a successful predictive maintenance system. Factors such as system scalability, real-time processing, and integration with existing maintenance workflows will help overcome challenges in real-world applications.

When sufficient real-world data is available, training deep learning models such as LSTM, RNN, CNN,

and GRU, or utilizing pre-trained models, may increase predictive accuracy. Ensemble techniques like voting, stacking, and bagging could further enhance model generalizability. Additionally, using techniques such as K-Nearest Neighbors (KNN) and Multiple Imputation by Chained Equations (MICE) alongside statistical methods for handling missing data could positively impact model performance.

To identify relevant variables and build the ideal graph structure, future studies could benefit from heuristic optimization algorithms such as genetic algorithms, hill climbing, ant colony, and gray wolf optimization, which are expected to improve predictive accuracy. Furthermore, a graph-based deep learning model (GNN - Graph Neural Network) could be trained on raw CAN Bus data, allowing the development of an integrated maintenance concept that considers both individual buses and the entire fleet. In addition, dynamic graph analysis and temporal graph networks could be explored to handle time-varying datasets. Hierarchical graph structures could be used to uncover relationships at different levels.

To detect unexpected situations in systems, recent algorithms such as Isolation Forest, One-Class SVM, autoencoders, and GANs could be tested alongside traditional statistical analyses. Moreover, reinforcement learning algorithms like Q-learning and Deep Q-Networks could replace supervised predictive maintenance models, enabling the optimization of maintenance strategies and improving vehicle performance. Transfer learning between different bus fleets could also be used to increase the generalizability of models, making them more adaptable to varying operational environments.

## References

Ambriško, Ľ., & Teplická, K. (2021). Proactive Maintenance as a Tool of Optimization for Vehicle Fleets, in Terms of Economic and Technical Benefits. *Acta Polytechnica Hungarica*, pp. 235-249.

Analytics, A. (2024). *Challenges of predictive analytics and how to overcome them*. (Amplitude Analytics) Retrieved from https://amplitude.com/explore/analytics/what-predictive-analytics

Bradbury, S., & Carpizo, B. (2018). *Digitally enabled reliability: Beyond.* McKinsey&Company.

Cheng, F., Zhou, C., Liu, X., Wang, Q., Qiu, J., & Zhang, L. (2023, August). Graph-Based Feature Selection in Classification: Structure and Node Dynamic Mechanisms. *IEEE TRANSACTIONS ON EMERGING TOPICS IN COMPUTATIONAL INTELLIGENCE*, pp. 1314-1328.

Cinieri, S. (2019). *Predictive maintenance of industrial vehicles based on supervised machine learning techniques.* Torino: Politecnico Di Torino.

Das, A. K., Goswami, S., Chakrabarti, A., & Chakraborty, B. (2017, December 1). A new hybrid feature selection approach using feature association map for supervised and unsupervised classification. *Expert Systems With Applications*, pp. 81-94.

Diez-Olivan, A., Ser, J. D., Galar, D., & Sierra, B. (n.d.). Data fusion and machine learning for industrial prognosis: Trends and perspectives towards Industry 4.0. *Information Fusion*.

Errandonea, I., & Arrizabalaga, S. (2020). Digital Twin for maintenance: A literature review. *Computers in Industry*.

Gbadamosı, A.-Q. O. (2023). *An Internet Of Things Enabled System For Real-Time.* University of the West of England, Bristol.

Hossan, M. S., & Chowdhury, B. (2019). Data-Driven Fault Location Scheme for Advanced Distribution Management Systems. *IEEE*.

Hosseini, S. (2021). A Comprehensive IoT-Enabled Predictive Maintenance Framework A Case Study of Predictive Maintenance for a Printing Machine. *Mathematics and Computer Science*.

Klathae, V. (2019). The Predictable Maintenance 4.0 by Applying Digital Technology: A Case Study of Heavy Construction Machinery. *Review of Integrative Business and Economics Research, Vol. 8*.

Lede, T. (2024, February 26). *Exploratory Data Analytics: Unveiling Insights for Data Enthusiasts*. (medium.com) Retrieved from https://medium.com/@thomas.lede.21/exploratory-data-analytics-unveiling-insights-for-data-enthusiasts-3aea84ae3c86

Martins, S. (2023, December 4). *Exploring The Benefits And Limitations Of Automated Machine Learning*. (Gradient Insight) Retrieved from https://gradientinsight.com/exploring-the-benefits-and-limitations-of-automated-machine-learning/

Mobley, R. K. (2002). *An Introduction to Predictive Maintenance.* Butterworth-Heinemann (Elsevier Science).

Nacchia, M., Fruggiero, F., Lambiase, A., & Bruton, K. (2021, March 12). A systematic mapping of the advancing use of machine learning techniques for predictive maintenance in the manufacturing sector. *Applied Industrial Technologies*.

Raposo, H., Farinha, J. T., Ferreira, L., & Galar, D. (2017). An integrated econometric model for bus replacement and determination of reserve fleet size based on predictive maintenance. *Eksploatacja i Niezawodnosc – Maintenance and Reliability*, pp. 358-368.

Saha, D., Mukherjee, S., Ankit, A., & Jadhav, A. (2024, April). Cognitive Diagnostics in Automotive using ML. *2024 IEEE International Students' Conference on Electrical, Electronics and Computer Science (SCEECS)*. Bhopal, India.

Scaife, A. D. (2024). Improve predictive maintenance through the application of artificial intelligence: A systematic review. *Results in Engineering, Volume 21*.

Schroeder, D. T., Styp-Rekowski, K., Schmidt, F., Acker, A., & Kao, O. (2019). Graph-based Feature Selection Filter Utilizing Maximal Cliques. *Sixth International Conference on Social Networks Analysis, Management and Security (SNAMS)*. Milan.

Srimrudhula. (2024, June 3). *What are the top 10 challenges in data mining, and how do you overcome them?* (medium.com) Retrieved from https://medium.com/@srimrudhula786/what-are-the-top-10-challenges-in-data-mining-and-how-do-you-overcome-them-91754aadf168

Theissler, A., Pérez-Velázquez, J., Kettelgerdes, M., & Elger, G. (2021, November). Predictive maintenance enabled by machine learning: Use cases and challenges in the automotive industry. *Reliability Engineering & System Safety_Elsevier*.

Turnbull, A. (2021). *Applications of machine learning in diagnostics and prognostics of wind turbine high speed generator failure*. University of Strathclyde.

Vic Barnett, T. L. (1978). Outliers In More Structured Situations. In *Outliers in Statistical Data*.

Victor, V. S., Schmeißer, A., Leitte, H., & Gramsch, S. (2024, April 15). Machine learning-based optimization workflow of the homogeneity of spunbond nonwovens with human validation. *ArXiv*.

Villar, E. F. (2024). *Machine Learning Maintenance Costs Prediction Model for Heavy-duty Alternative Fuel and Diesel Vehicles*. West Virginia University.

Zhang, Z. (2012). *Feature Selection from Higher Order Correlations*. York: University of York.

Zhao, Y., Li, T., & Zhang, X. (2019). Artificial intelligence-based fault detection and diagnosis methods for building energy systems: Advantages, challenges and the future. *Renewable and Sustainable Energy Reviews, 109*.